\documentclass[journal]{IEEEtran}

\usepackage[pdfborder=0 0 0,colorlinks=true,urlcolor=blue,citecolor=blue, linkcolor=blue]{hyperref}

\ifCLASSINFOpdf
  \usepackage[pdftex]{graphicx}
\else
\fi

\usepackage{url}

\newcommand{\TDPvenue}{2025~}

\begin{document}

\title{RoboCup Rescue \TDPvenue Team Description Paper \\
 UruBots}

\author{Kevin~Farias,
        Pablo~Moraes,
        Igor Nunes,
        Juan Deniz,
        Sebastian Barcelona,
        Hiago Sodre, 
        William Moraes,
        Mónica Rodriguez,
        Ahilén Mazondo,
        Vincent Sandin,
        Gabriel da Silva,
        Victoria Saravia,
        Vinicio Melgar,
        Santiago Fernandez,
        Ricardo Grando
\thanks{Robo Hero is with the Department
of Robotics Gurus, Moon University, Moon, e-mail: .}
\thanks{J. Doe and J. Doe are with Anonymous University.}
}

\markboth{RoboCup Rescue \TDPvenue TDP Collection}%
{Hero \MakeLowercase{\textit{et al.}}: UruBots}

\maketitle

\begin{flushleft}
\textbf{Info}\\
\hspace{10pt} Team Name: \hfill UruBots\\
\hspace{10pt} Team Institution: \hfill Technological University of Uruguay\\
\hspace{10pt} Team Country: \hfill Uruguay\\
\hspace{10pt} Team Leader: \hfill Kevin Farias\\
\hspace{10pt} Team URL: \hfill \url{urubots.uy}
\\
\vspace{5pt}
\hspace{10pt} Qualification Video: 
 \url{https://youtu.be/uDzRcIA3kGk}
\end{flushleft}

\begin{abstract}
This paper describes the approach used by Team UruBots for participation in the 2025 RoboCup Rescue Robot League competition. Our team aims to participate for the first time in this competition at RoboCup, using experience learned from previous competitions and research. 

We present our vehicle and our approach to tackle the task of detecting and finding victims in search and rescue environments. Our approach contains known topics in robotics, such as ROS, SLAM, Human Robot Interaction and segmentation and perception. Our proposed approach is open source, available to the RoboCup Rescue community, where we aim to learn and contribute to the league.
\end{abstract}


\IEEEpeerreviewmaketitle

\section{Introduction}

\IEEEPARstart{T}{eam} UruBots was established in 2022 at the Technological University of Uruguay, campus of Rivera. The team has attended the FIRA Roboworld Cup competitions of 2023 and 2024, where first place was achieved in the Mission Impossible league in 2023 and third place in both the Autonomous Car and Humanoids Leagues in 2024. We aim to attend the Robocup competition for the first time, specifically in the Rescue League as it relates to our previous experience both in competitions \cite{sodre2024urubots, moraes2024urubots} and in research \cite{deniz2024real, moraes2024behavior, grando2024improving}. 

\begin{figure}[!t]
\centering
\includegraphics[width=\linewidth]{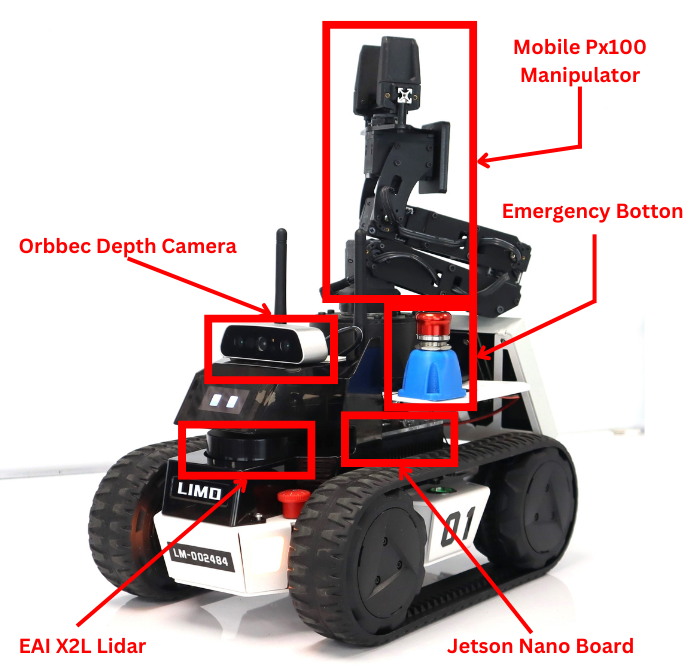}
\caption{Limo Robot for Competition Configuration.}
\label{fig:robotPhoto}
\end{figure}

Our proposed robotics system is based on a mobile robot equipped with a differential tracked locomotion system, which allows movement both in flat and non-flat terrains. We equipped our mobile robot with a robotic arm and a depth camera to perform manipulation-related tasks and to perform perception and segmentation of objects in a rescue-related scenario. We also created a software system based on a Robotics Operating System with a mission planner that combines different packages of mapping, navigation, manipulation, and perception to perform rescue-related tasks. Overall, our proposed robotic solution for the RoboCup Rescue League 2025 can be seen in Figure \ref{fig:robotPhoto}.

\subsection{Scientific Publications}

Since 2022, our team has been developing research in topics related to rescue tasks and environments \cite{deniz2024real, moraes2024behavior, grando2024improving}. In Deniz et al. \cite{deniz2024real}, we proposed a Real-time Robotic Situational Awareness for Accident Prevention in Industry. We proposed a system based on the mobile robot LoCobot adapted to be able to perform human-robot interaction tasks in order to be able to communicate with the end user and detect dangerous situations in an industrial simulated scenario. We also created a ROS-based package to perform autonomous navigation in a simulated industrial environment, combining information with the object detection performed by the YOLO network to engage in interaction with a person in the scenario. We concluded that our system was able to detect situations such as when a person has no helmet or vest and alert the user. Although focused on the industrial context, this work relates to rescue tasks once autonomous navigation and object detection are some of the demands in both contexts.

In Moraes et al. \cite{moraes2024behavior}, we performed an ablation study of convolutional neural networks to perform the autonomous navigation of a mini-autonomous car through behavior cloning. We showed that by adding or removing layers, it was possible to achieve better navigation in some situations. This work provided us with expertise in neural networks and in autonomous navigation. In Grando et al. \cite{grando2024improving} it was shown that Reinforcement Learning can also be used to improve the autonomous navigation of mobile robots. Specifically, this work showed that by delaying the learning in an RL agent, it was possible to achieve good performance in goal-oriented tasks in known and unknown scenarios. 

We also developed some works in local events  \cite{sodre2023aplicacion, pablo2024intera} related to human-robot interaction and object detection. We also highlight some previous TDPs in other competitions related to mini autonomous car \cite{moraes2024urubots} and unmanned aerial vehicles \cite{sodre2024urubots}, where in the first\cite{moraes2024urubots} we showed an approach based on machine learning for autonomous car navigation and approach based on visual odometry to perform navigation-related tasks in the second\cite{sodre2024urubots}. 

\section{Hardware Description}

This section will present the proposed hardware description for the RoboCup 2025 Rescue competition. Our mechanical and hardware setup was done with the objective of being able to tackle tasks of the competitions and to have space for adaptations in case of necessity during the competition week. The sections itemized below and described as follows:

\begin{itemize}
 \item Robotic Plataform
 \item Robotic Manipulator 
 \item Processing and Communication platform
 \item Vision Plataform
\end{itemize}

\subsection{Robotic Plataform}

The robotic platform system is based on a Robot Limo \cite{GenerationRobots-Limo}. This robot was created by the company AgileX and has an integration of hardware with a mobile robotic base, a 2d lidar, a depth camera, a dedicated processing board, and other components that allow for SLAM. In Table \ref{tab:SystemRobot1} below, it is possible to view the description of the components and technical details of our platform.

\begin{table}[h]
\renewcommand{\arraystretch}{1}
 \tabcolsep=0.1cm
\caption{Technical specifications of our robotic platform}
\label{tab:SystemRobot1}
\centering
\begin{tabular}{|l|r|}
\hline
Attribute & Value \\ \hline
Name & Limo  \\
Overall dimension  & 322*220*251mm \\
Wheelbase & 200mm \\
Tread  & 175mm \\
Dead load  & 4.8kg \\
Payload & 4kg \\
Minimum ground clearance & 24mm \\
Drive type & Hub motor(4x14.4W) \\
No-load max. speed  & 1m/s \\
LiDAR & EAI X2L\\
Depth Camera & ORBBEC Dabai\\
IPC & Jeston Nano\\
Battery  & 5.2AH 12V\\
Operating System  & Ubuntu 18.04\\
ROS Version & ROS1 Melodic\\
App Control Range & 10m\\
Control Method & Mobile App/Command Control\\

\hline
\end{tabular}
\end{table}

\begin{figure}[h]
\centering
\includegraphics[scale=.35]{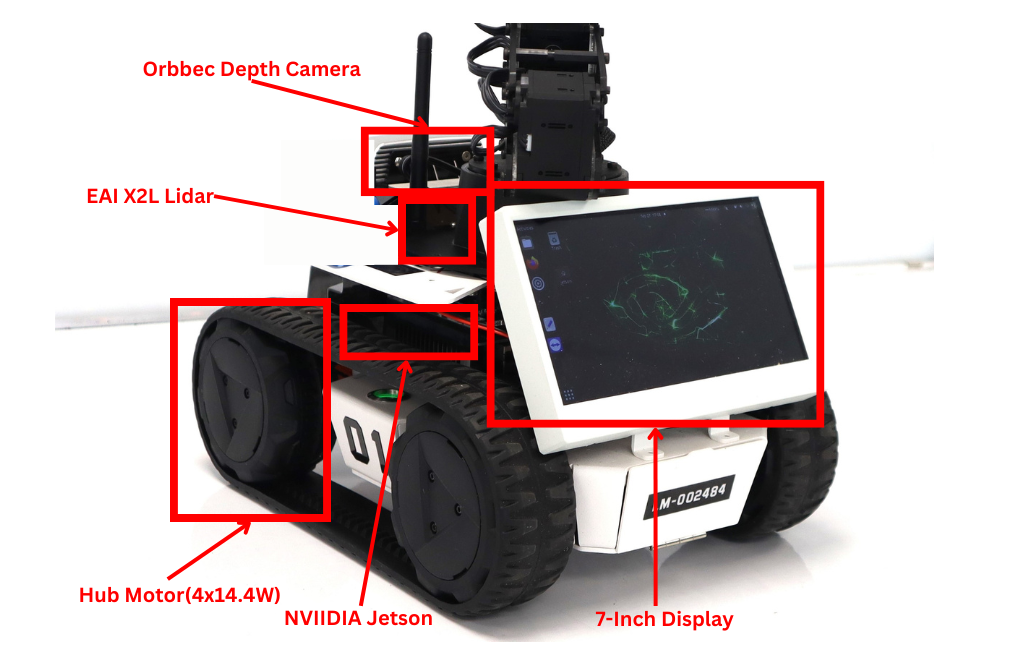}
\caption{LIMO Mobile Robot: Standard Version.}
\label{fig:limo1}
\end{figure}

Our version of the LIMO mobile robot includes a Nvidia Jetson Nano, an EAI X2L LiDAR, and an Orbbec DaBai depth camera. These equipments give the robot the possibility of a robust perception of its environment \cite{GenerationRobots-Limo}. It is ideal for developing applications for autonomous navigation, obstacle avoidance, and visual recognition. The components of our version of our LIMO used can be seen in Figure \ref{fig:limo1}, highlighting the red color.

\subsection{Robotic Manipulator}

\begin{figure}[h]
\centering
\includegraphics[scale=.35]{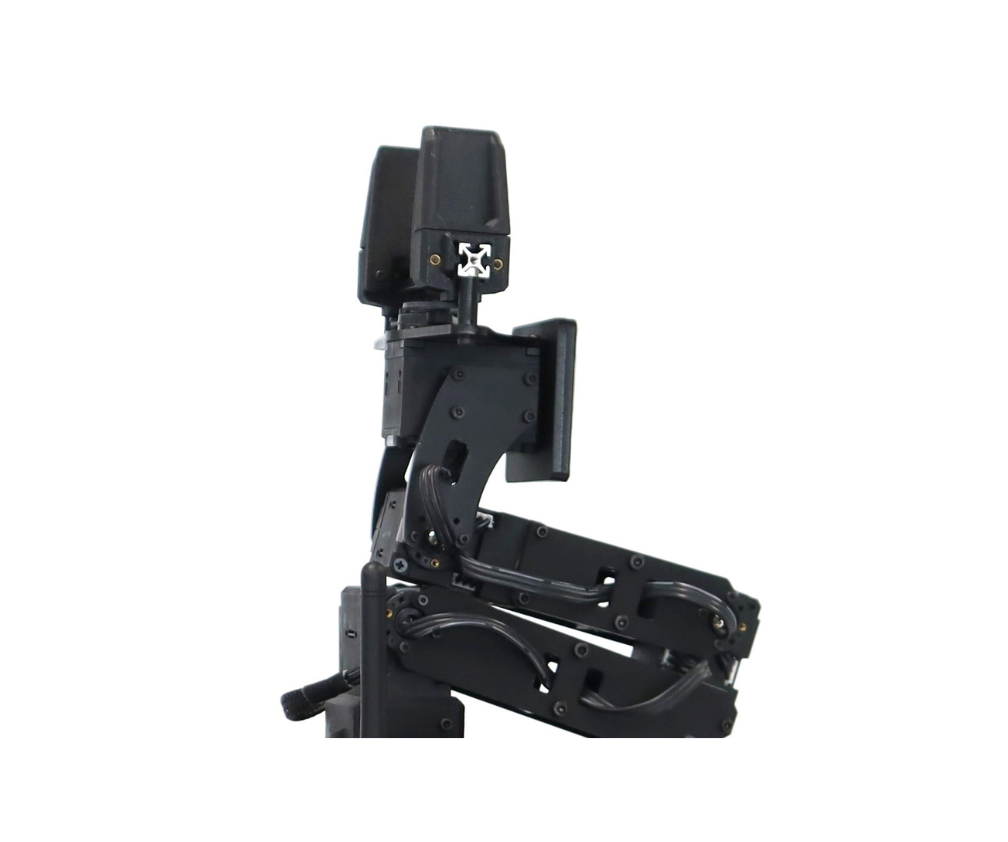}
\caption{PincherX-100 Manipulator.}
\label{fig:mani}
\end{figure}

The Manipulator added to the Robot LIMO system is a PincherX-100 manipulator from the company Trossen Robotics. The PincherX-100 Robot Arm uses Dynamixel X-Series servo motors, which offer high torque in a small form factor\cite{PincherX100-Manipulator}, suiting for this particular application. In Figure \ref{fig:mani}, it's possible to see an image of this version of Manipulator.

This Manipulator enables access to Dynamixel Wizard software as well as ROS via the Robotis Dynamixel U2D2. The PincherX-100 offers 4 degrees of freedom and a full 360 degree of rotation. It has a range of 300mm, repeatability of 5mm, accuracy of 8mm and a useful working load of 50g. What makes it useful and applicable to the robotic platform being used. 

\subsection{Processing and Communication platform}

\begin{figure}[h]
\centering
\includegraphics[scale=.25]{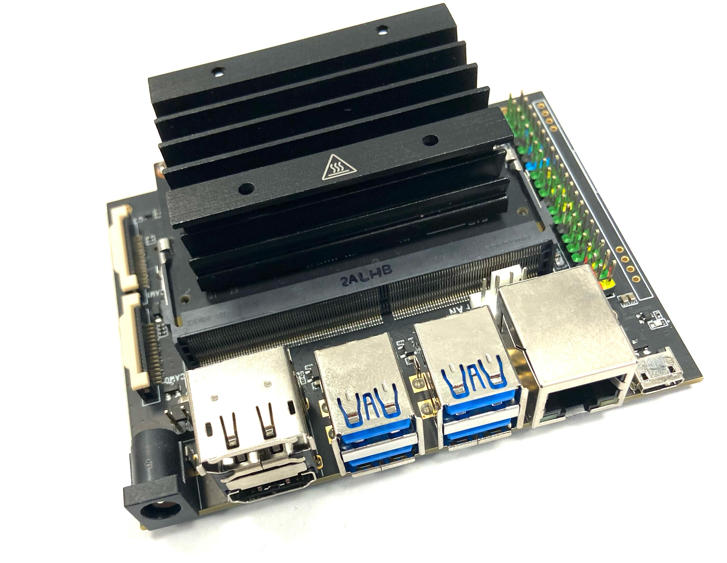}
\caption{NVIDIA Jetson Nano.}
\label{fig:jetson}
\end{figure}

This model of the LIMO robot integrates an NVIDIA Jetson NANO used for processing and controlling our robotics solutions. It is a compact platform with a 128-core CUDA GPU based on the Maxwell architecture. For storage, it uses a microSD card, although the B01 version provides NVMe support through the PCIe M.2 port, which significantly increases data access speed. Its 128 CUDA cores GPU and compatibility can be integrated with sensors such as LiDAR, cameras, and encoders, facilitating environmental awareness and route planning. In addition, it has low energy consumption and supports deep learning frameworks such as TensorFlow and PyTorch, Figure \ref{fig:jetson} presents an image of NVIDIA Jetson Nano used.

\begin{figure}[h]
\centering
\includegraphics[scale=.55]{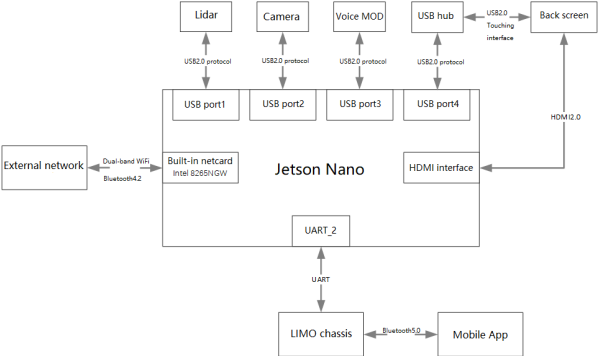}
\caption{Communication topology of the Limo Robot. }
\label{fig:communication}
\end{figure}

The LIMO chassis incorporates a Bluetooth 5.0 module that allows remote control and connection with a mobile application. It connects directly to the Nano board through a UART interface, allowing it to be controlled. In addition, it has a USB HUB that offers two USB ports and one Type-C port, all compatible with the USB 2.0 protocol. The rear screen, with a touch function, is connected to the USB HUB via the USB 2.0 interface. Figure \ref{fig:communication} shows the communication topology of the processing board. 

\newpage

\begin{figure}[h]
\centering
\includegraphics[scale=.45]{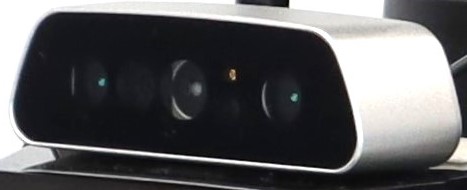}
\caption{ORBBEC Dabai Deep Camera.}
\label{fig:camera}
\end{figure}

The LIMO robot also integrates a camera that allows image detection and segmentation for a robotic platform of this type. The camera in question is the ORBBEC Dabai depth camera, presented in Figure \ref{fig:camera}, which is a depth camera based on binocular structured light 3D imaging technology. 

Its image quality offers resolutions of 640x400 pixels at 30FPS and 320x200 pixels at 30FPS. For color images, it reaches 1920x1080 pixels at 30FPS. Its accuracy is 6 mm at a distance of 1 meter, with 81 percent of the viewing area contributing to accuracy calculations. Its depth range covers 0.3 to 3 meters, allowing accurate captures of various
applications.

\section{Software}

Our proposed software system is based on the Robotics Operating System, combining both existing and developed packages. Every part of our software system is presented below.

\subsection{SLAM}

Our proposed Simultaneous Localization and Mapping (SLAM) is based on the Google Cartographer package for ROS \cite{hess2016real}. The SLAM system gets information for a 2D Lidar and the robots' odometry, where a map of a scenario was manually generated, as can be seen in Figure \ref{fig:map}

\begin{figure}[h]
\centering
\includegraphics[scale=.35]{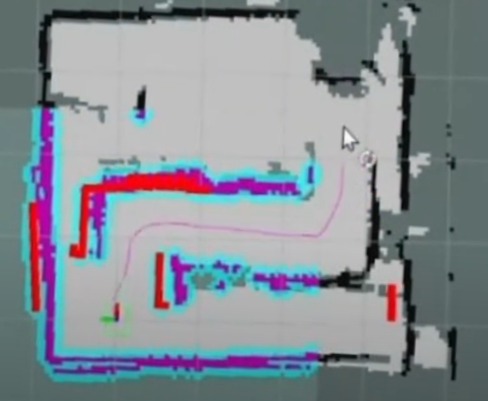}
\caption{Mapping of a scenario.}
\label{fig:map}
\end{figure}

We used the robot's package\footnote{$https://github.com/agilexrobotics/limo_ros$} to manually create the map and use it as input for the robots' SLAM. As can be seen in Figure \ref{fig:map}, a target position can be defined through the RVIZ interface and also using the robots' target position topic. Figure \ref{fig:map} shows the robot's frame and then the calculated path toward a manually set target on the RVIZ interface.

\subsection{Object Detection and Manipulation}

\begin{figure}[h]
\centering
\includegraphics[scale=.35]{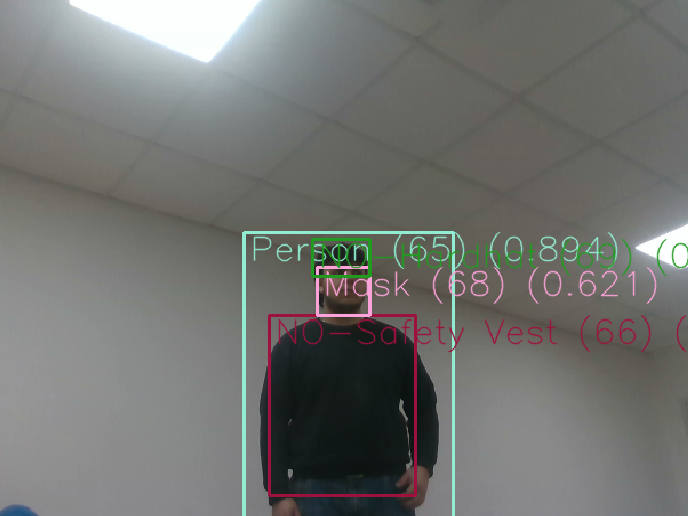}
\caption{Detection of a person using YOLO framework.}
\label{fig:persondetection}
\end{figure}

Our detection of objects and people is based on our previous research work \cite{deniz2024real}, where a YOLO model was trained to detect a person in a dangerous situation. The trained model was based on the YOLOv5 framework and its package ROS \footnote{$https://github.com/mgonzs13/yolo_ros$} was possible to detect a person and objects in the scenario. Figure \ref{fig:persondetection} shows a sample of detection using our trained model, which is capable of detecting a person and also advising if the person has a helmet and a protective vest. For the competition, we aim to train a new version of the model, taking into consideration objects for the tasks demanded at the competition.

The proposed manipulation system is based on the PincherX-100 manipulator and its ROS package\footnote{$github.com/Interbotix/interbotix_ros_manipulators$}. The package provides ROS topics and services that allow the set of a position and orientation for the arm's end effector, which are set taking values from a transformation from the YOLO detection's outputs. 

\section{Qualification Experiment}

According to the competition's theme of rescue tasks, a scenario was developed where the robot should navigate autonomously to a specific desired point. For that, in Figure \ref{fig:map} below, it is possible to visualize the comparison of the scenario built for the mapping and navigation tests with the robot and the map generated by mapping with the robot.

\begin{figure}[h]
\centering
\includegraphics[scale=.2]{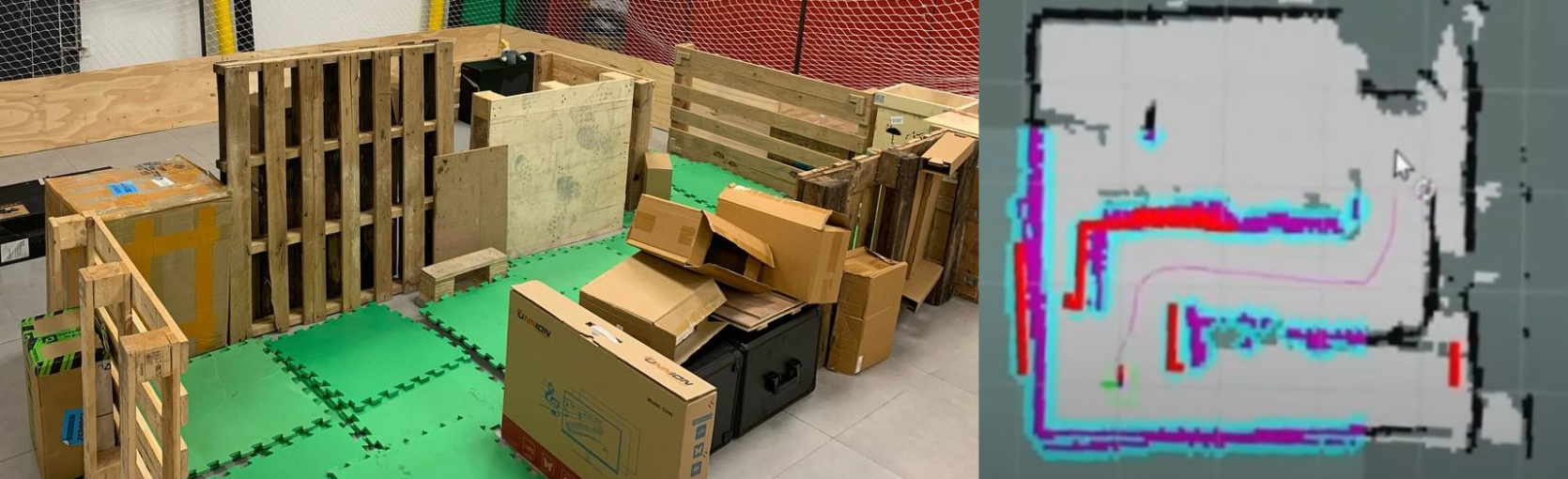}
\caption{Experiment Scenario and Generated Map.}
\label{fig:map}
\end{figure}

With the generated map, the team developed some navigation tests with the presented robot, challenging its speed and ability to navigate autonomously in the proposed scenario. The experiment proposal consists of the following steps:

\begin{itemize}
 \item Place the robot in the starting position
 \item Open the generated map in Rviz
 \item Set the navigation goal
 \item Monitor the robot during navigation
\end{itemize}

\begin{figure}[h]
\centering
\includegraphics[scale=.18]{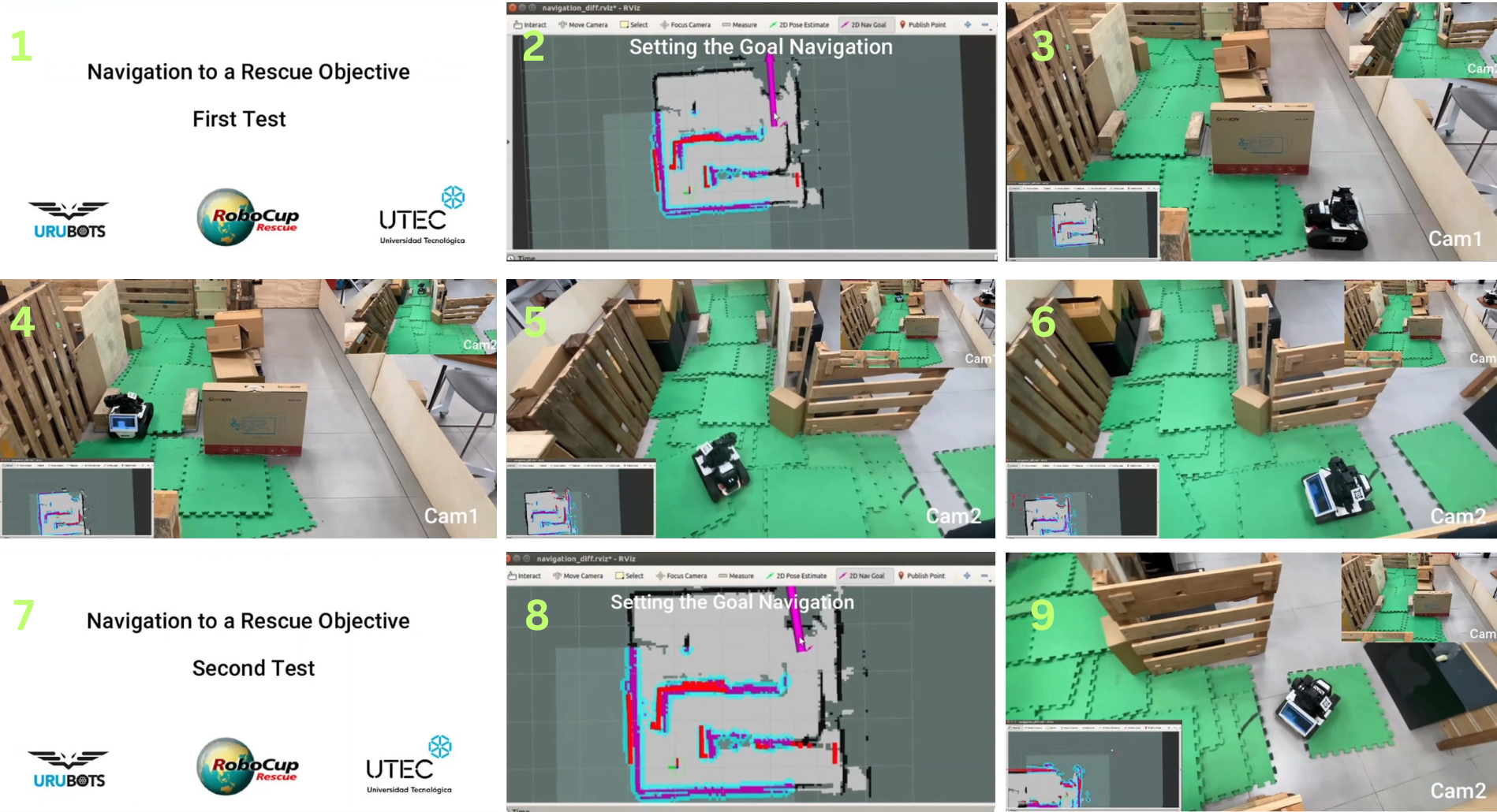}
\caption{Timeframe of our qualification experiment.}
\label{fig:timeframe}
\end{figure}

In accordance with this organization of experiments, some recompiled tests were carried out and presented in the Qualification video \footnote{$https://youtu.be/gEDb6GOWuMo$}. We carried out two main tests of different distances at different speeds, where the robot would have to cross the maze autonomously and also pass through floor-leveling obstacles and a ground platform that makes it difficult because it is steep. In Figure \ref{fig:timeframe}, it also presented the timeframe of the video qualification. \cite{UruBots-Video}

As shown in Figure \ref{fig:timeframe}, the training scenario was made using multiple items, such as wooden and cardboard boxes, wooden pallets, and layers of foam mat tiles. The training ground tried to be as similar as possible to previously used ones in the actual competitions, although due to space and material limitations, it was not possible to replicate the exact conditions, mainly the highly steep slopes and obstacles.

\section{Conclusion}

In this team description paper, we provide the first approach of our team for the Robocup Rescue League. We focus on bringing together existing hardware and software to provide a reliable solution to attend the RoboCup 2025 competition. Through the experiments performed and the methodology presented we aim to attend the competition for the first time, looking forward to learn and attend future editions.

\appendices

\section{Team members and Their Contributions}

Many students and researchers at the Technological University of Uruguay contribute to the team. The following list as it appears in the author's list:

\begin{itemize}
  \item  {Kevin Farias} \hfill Hardware Setup
  \item  {Pablo Moraes} \hfill Mechanical Setup
  \item  {Igor Nunes}   \hfill Manipulation Setup
  \item  {Juan Deniz}   \hfill Software Development
  \item  {Sebastian Barcelona} \hfill Software Development
  \item  {Hiago Sodre} \hfill  Mechanical Setup
  \item  {William Morae} \hfill Software Development
  \item  {Mónica Rodriguez} \hfill Software Development
  \item  {Ahilén Mazondo} \hfill Mechanical Setup
  \item  {Vincent Sandin} \hfill Manipulation Setup
  \item  {Gabriel da Silva} \hfill Mechanical Setup
  \item  {Victoria Saravia} \hfill Mechanical Setup
  \item  {Vinicio Melgar} \hfill Mechanical Setup
  \item  {Santiago Fernandez} \hfill Mechanical Setup
  \item  {Ricardo Grando} \hfill Software Development
\end{itemize}

\section*{Acknowledgment}

The authors would like to thank the Technological University of Uruguay for the continued support to attend competitions of robotics provided to our team.

\ifCLASSOPTIONcaptionsoff
  \newpage
\fi

\bibliographystyle{IEEEtran}
\bibliography{references}

\end{document}